\documentclass[runningheads]{llncs}
\usepackage[T1]{fontenc}
\usepackage{graphicx,verbatim}
\usepackage{amsfonts}
\usepackage{amssymb}
\usepackage{algorithm}
\usepackage{algpseudocode}

\usepackage{enumitem}
\usepackage{booktabs}
\usepackage{pifont}

\usepackage{xcolor}
\definecolor{ourscolor}{RGB}{232,245,233}
\usepackage{pifont}
\usepackage[most]{tcolorbox}
\definecolor{highlightgreen}{RGB}{34,120,69}
\definecolor{ourscolor}{RGB}{232,245,233}
\definecolor{darkgreen}{rgb}{0.0,0.5,0.3}
\definecolor{algproto}{HTML}{1D4ED8}  
\definecolor{algfuse}{HTML}{B45309}   
\definecolor{alggate}{HTML}{15803D}   
\usepackage[noEnd=true, indLines=true]{algpseudocodex}

\definecolor{hlprotoC}{HTML}{DBEAFE}  
\definecolor{hlfuseC}{HTML}{FEF3C7}   
\definecolor{hlgateC}{HTML}{DCFCE7}   

\definecolor{gold}{HTML}{FFF3CD}
\definecolor{silver}{HTML}{E8E8E8}
\definecolor{bronze}{HTML}{F5E6D3}
\definecolor{protonrow}{HTML}{EBF0FF}
\definecolor{deltagreen}{HTML}{16A34A}
\definecolor{deltared}{HTML}{DC2626}
\definecolor{sepgray}{HTML}{CCCCCC}

\usepackage{booktabs}
\usepackage{tabularx}
\usepackage{multirow}
\usepackage{colortbl}
\usepackage{makecell}
\usepackage{siunitx}
\usepackage{caption}
\usepackage{hyperref}
\usepackage{tikz}

\begin{document}
\renewcommand{\vec}{\@ifstar{\vec@star}{\vec@}}
\title{PROTON: Prototype-Based Test-Time Online OOD Detection for Medical VLMs}
\titlerunning{PROTON}


\author{Abhijit Das$^1$,
Nichula Wasalathilaka$^2$,
Yifan Lu$^1$,
Adinath Dukre$^1$,
Dwarikanath Mahapatra$^{3}$,
Shadab Khan$^4$, 
Imran Razzak$^{1,5}$}
\authorrunning{Das et al.}
\institute{$^1$ MBZUAI, Abu Dhabi, UAE,
    $^2$ University of Peradeniya, Sri Lanka,\\
    $^3$ Khalifa University, Abu Dhabi, UAE
    $^4$ ADIA Lab, Abu Dhabi, UAE,\\
    $^5$ MedOS, Abu Dhabi, UAE}
  
\maketitle              

\begin{abstract}
    Medical vision-language models (VLMs) enable zero-shot clinical image classification, yet reliably detecting out-of-distribution (OOD) inputs at deployment remains an open problem. No static scoring method works across all shift types: Maximum Concept Matching (MCM) on FLAIR achieves 76.4\% AUROC for far-OOD but only 42.4\% for covariate shifts such as ultra-wide-field fundus images \textemdash effectively random. We trace this to a structural mismatch: covariate-shifted inputs are indistinguishable from in-distribution samples in softmax space, yet occupy distinct regions in the VLM's embedding space. To exploit this untapped signal, we propose \textbf{PROTON} (\textbf{PRO}totype-based \textbf{T}est-time \textbf{ON}line OOD detection), a lightweight post-hoc module that maintains an online \textbf{prototype bank} from high-confidence test predictions and adaptively fuses prototype distance with MCM scoring via stream-level variance statistics, requiring no model modification, training data, or prompt engineering. On the ophthalmology benchmark (FLAIR + FIVES), PROTON improves MCM by \textbf{+23.9 AUROC on covariate shift}, +8.8 on semantic, and +8.1 on far-OOD \textemdash the only zero-shot method to improve all three without hierarchical prompts or labeled data. Code is online.\footnote{\href{https://github.com/GenMI-Lab/PROTON}{github.com/GenMI-Lab/PROTON}, \quad Also visit the project page: \href{https://genmi-lab.github.io/PROTON}{PROTON}.}

\end{abstract}

\section{Introduction}

    Medical vision-language models (VLMs) such as FLAIR~\cite{Silva_Rodr_guez_2025}, UniMedCLIP~\cite{khattak2024unimed}, and QuiltNet~\cite{ikezogwo2023quilt} enable zero-shot clinical classification, yet deployed systems inevitably encounter out-of-distribution (OOD) inputs---misrouted pathologies, novel imaging devices, or non-clinical submissions---risking silent misdiagnosis as VLM-based screening scales. Existing OOD detectors for VLMs are \emph{static}: MCM~\cite{ming2022delving}, energy~\cite{liu2020energy}, and GL-MCM~\cite{miyai2025gl} apply a fixed scoring function regardless of deployment context. Ju et al.~\cite{ju2025delving} show that no static method works across all shift types on medical VLMs, a finding corroborated across 14 datasets by OpenMIBOOD~\cite{gutbrod2025openmibood}. Recent efforts address detection and adaptation separately: HVL~\cite{lai2025hierarchical} and GLAli~\cite{YanJie_Global_MICCAI2025} improve medical OOD detection but require labeled ID data; BCA+~\cite{li2025generalizing}, SCA~\cite{zhou2025self}, and UL-TTA~\cite{kim2025ultra} adapt VLMs at test time but assume a closed label set, aiming to \emph{accept} shifted inputs rather than flag them; and OODD~\cite{linoodbench} applies test-time detection via a dynamic dictionary but requires ID training data and targets only standard classifiers on CIFAR or ImageNet.
    
    \begin{figure}[t]
        \centering
        \includegraphics[width=0.8\textwidth]{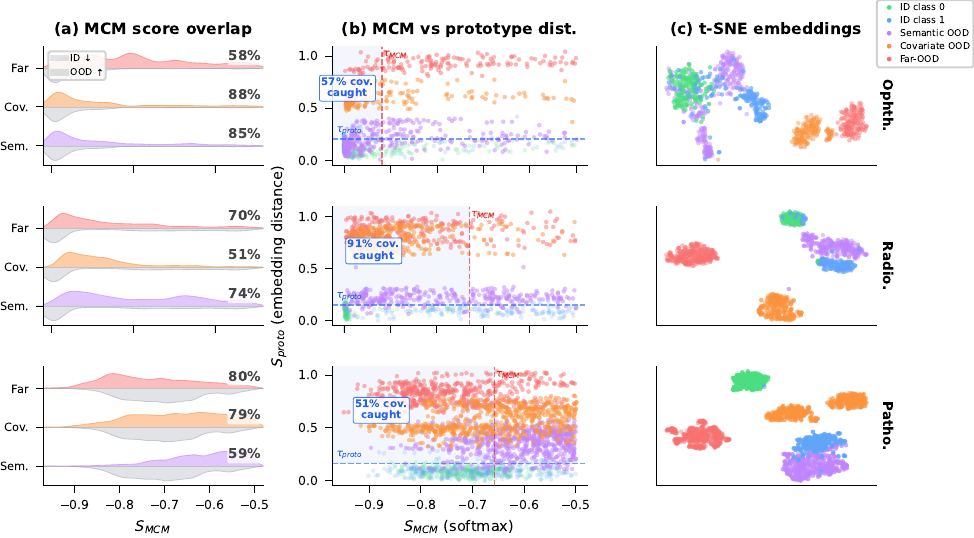}
        \caption{\textbf{MCM's blind spot.}
        \textbf{(a)}~MCM score overlap between ID and OOD (51--88\% across domains).
        \textbf{(b)}~Prototype distance ($y$-axis) separates covariate samples that MCM ($x$-axis) cannot; blue zone: 51--91\% of covariate OOD caught by PROTON. 
        \textbf{(c)}~t-SNE confirms geometric separation that softmax collapses.
        Rows (Top to Bottom, in order): FLAIR, UniMedCLIP, QuiltNet.}
        \label{fig:intro}
    \end{figure}
    
    The common thread across these failures is a missed signal: covariate-shifted samples produce MCM scores indistinguishable from ID yet occupy \emph{distinct regions} in the VLM's embedding space (Fig.~\ref{fig:intro}). Static methods collapse this geometry into a scalar via softmax, discarding exactly the separation that matters. Building on this insight, we propose \textbf{PROTON} (\textbf{PRO}totype-based \textbf{T}est-time \textbf{ON}line OOD detection), a lightweight post-hoc module atop a frozen VLM. PROTON maintains an \textbf{online prototype bank} of per-class centroids built from high-confidence test predictions, computes a \emph{prototype distance score} via cosine similarity to the nearest centroid, and \emph{adaptively fuses} this with MCM scoring using a weight derived from running MCM variance via \emph{Welford's online algorithm}---recovering the covariate separation that softmax discards. Concretely, we make the following \textbf{contributions:}
    
    \begin{enumerate}[leftmargin=*,itemsep=1pt,topsep=1pt,parsep=0pt,partopsep=0pt]
        \item We formalize the unexploited gap between test-time adaptation and static OOD detection in medical VLMs, and propose PROTON---to our knowledge, the first model-agnostic, prompt-free framework bridging both via an online prototype bank, requiring no training data or model modification.
        \item We introduce variance-driven adaptive fusion that reweights softmax and embedding-based scores via Welford's online algorithm, automatically upweighting prototype distance when MCM becomes unreliable (covariate shift) while preserving its strength on far-OOD without any shift-type supervision.
        \item PROTON is the only zero-shot method to improve all three shift types on the ophthalmology benchmark (largest gain on covariate, \textbf{+23.9} AUROC), and the effect is not MCM-specific---it lifts six base scores and transfers to radiology and pathology, robust to low ID stream fractions at ${\sim}0.4$\,MB.
    \end{enumerate}

    \begin{figure*}[t]
        \centering
        \includegraphics[width=\textwidth]{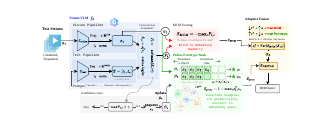}
        \caption{\textbf{Overview of PROTON.} A frozen VLM produces embedding $\mathbf{e}_t$ and softmax probabilities $\mathbf{p}_t$ per test image. $S_{\mathrm{MCM}}$ scores softmax confidence; $S_{\mathrm{proto}}$ measures cosine distance to online class prototypes in per-class FIFO queues. Adaptive fusion weights both via MCM stream variance ($\alpha_t$), and a confidence gate prevents OOD contamination of prototypes.}
    \label{fig:architecture}
    \end{figure*}

\section{Method}
\label{sec:method}

    Given the gap identified above, we now formalize the setting and describe PROTON's three components (Fig.~\ref{fig:architecture}): (1)~an \emph{online prototype bank} (Sec.~\ref{sec:proto_bank}), (2)~a \emph{prototype distance score} (Sec.~\ref{sec:proto_score}), and (3)~\emph{adaptive fusion} (Sec.~\ref{sec:fusion}); it adds no learnable parameters and degrades gracefully to pure MCM before prototypes accumulate.

    Let $f_\theta$ be a frozen medical VLM with vision encoder $f_v$ and text encoder $f_t$, mapping inputs into a shared $d$-dimensional space ($d{=}512$ for FLAIR). Text embeddings $\mathbf{T} = [\mathbf{t}_1, \ldots, \mathbf{t}_C] \in \mathbb{R}^{C \times d}$ for $C$ in-distribution class names are computed once and cached. For each test image $x_t$, the vision encoder produces $\ell_2$-normalized embedding $\mathbf{e}_t = f_v(x_t) \in \mathbb{R}^d$, yielding softmax probabilities $\mathbf{p}_t = \mathrm{softmax}(\mathbf{e}_t^\top \mathbf{T} / \tau)$ and the MCM score~\cite{ming2022delving} $S_{\mathrm{MCM}}(x_t) = -\max_c\, p_{t,c}$ (higher $=$ more OOD). Images arrive as an unlabeled stream; the detector scores each $x_t$ using $f_\theta$ and past observations $x_1, \ldots, x_{t-1}$, with no training data, calibration labels, or weight updates permitted.

    \subsection{Online Prototype Bank}
    \label{sec:proto_bank}
    
    For each class $c \in \{1, \ldots, C\}$, the bank maintains a FIFO queue $\mathcal{D}_c$ of maximum size $M$. After scoring $x_t$, a \emph{confidence gate} determines whether to update: if $\max_c\, p_{t,c} \geq \gamma$, embedding $\mathbf{e}_t$ is enqueued into $\mathcal{D}_{\hat{c}}$ ($\hat{c} = \arg\max_c\, p_{t,c}$); otherwise $x_t$ is scored but excluded from the bank, preventing low-confidence \textemdash potentially OOD \textemdash embeddings from contaminating prototypes. When $|\mathcal{D}_c|$ exceeds $M$, the oldest entry is evicted so prototypes track recent deployment statistics.
    
    The class prototype is the $\ell_2$-normalized mean of stored embeddings:
    \begin{equation}
        \boldsymbol{\mu}_c = \frac{\bar{\mathbf{e}}_c}{\|\bar{\mathbf{e}}_c\|_2}, \quad \bar{\mathbf{e}}_c = \frac{1}{|\mathcal{D}_c|} \sum_{\mathbf{e} \in \mathcal{D}_c} \mathbf{e}.
        \label{eq:prototype}
    \end{equation}
    Normalization ensures cosine similarity via dot product, consistent with the VLM's contrastive training. The detector is \emph{calibrated} once every class accumulates $K_{\min}$ samples ($K_{\min}{=}5$); before this, PROTON defaults to MCM.
    
    \subsection{Prototype Distance Score}
    \label{sec:proto_score}
    
    Once calibrated, PROTON scores each sample by its embedding-space proximity to the nearest prototype:
    \begin{equation}
        S_{\mathrm{proto}}(x_t) = 1 - \max_{c} \; \mathbf{e}_t^\top \boldsymbol{\mu}_c.
        \label{eq:proto_score}
    \end{equation}
    Since both vectors are $\ell_2$-normalized, this equals one minus the maximum cosine similarity: low for ID samples near their prototype, high for OOD samples distant from all prototypes---capturing the embedding-space separation that MCM's softmax collapses.

    \begin{algorithm}[t]
    \scriptsize
    \caption{PROTON: Online OOD Detection at Test Time}
    \label{alg:proton}
    \begin{algorithmic}[1]
    \Require Frozen VLM $f_\theta$, prompts $\{y_c\}_{c=1}^{C}$, hyperparams $\gamma, M, K_{\min}, \alpha_{\min}, \alpha_{\max}, \sigma^2_0$
    \State $\mathbf{T} \gets [f_t(y_1), \ldots, f_t(y_C)]$; \;\; $\mathcal{D}_c \gets \emptyset\,\forall c$; \;\; $(n,\bar{S},M_2)\gets 0$ \Comment{cache text; init bank+Welford}
    \For{each test image $x_t$ in stream}
        \State $\mathbf{e}_t \gets f_v(x_t)$; \;\; $\mathbf{p}_t \gets \mathrm{softmax}(\mathbf{e}_t^\top \mathbf{T}/\tau)$; \;\; $S_{\mathrm{MCM}} \gets -\max(\mathbf{p}_t)$
        \State update $(n,\bar{S},M_2)$ via Welford $\Rightarrow \sigma^2_t$ \Comment{\textcolor{algfuse}{$O(1)$ stream variance}}
        \If{$|\mathcal{D}_c| \geq K_{\min}\;\forall c$} \Comment{calibrated}
            \State \textcolor{algproto}{$S_{\mathrm{proto}} \gets 1 - \max_c\, \mathbf{e}_t^\top \boldsymbol{\mu}_c$} \Comment{\textcolor{algproto}{prototype distance}}
            \State \textcolor{algfuse}{$\alpha_t \gets \alpha_{\max} - \sigma\!\big(100(\sigma^2_t - \sigma^2_0)\big)(\alpha_{\max}-\alpha_{\min})$}
            \State \textcolor{algfuse}{$S_t \gets \alpha_t S_{\mathrm{MCM}} + (1-\alpha_t) S_{\mathrm{proto}}$} \Comment{\textcolor{algfuse}{adaptive fusion}}
        \Else
            \State $S_t \gets S_{\mathrm{MCM}}$ \Comment{pure-MCM fallback}
        \EndIf
        \If{\textcolor{alggate}{$\max(\mathbf{p}_t) \geq \gamma$}} \Comment{\textcolor{alggate}{confidence gate}}
            \State enqueue $\mathbf{e}_t \to \mathcal{D}_{\hat{c}}$ ($\hat{c}{=}\arg\max \mathbf{p}_t$), evict oldest if full; update $\boldsymbol{\mu}_{\hat{c}}$ \eqref{eq:prototype}
        \EndIf
        \State \textbf{emit} $S_t$ \Comment{per-sample OOD score}
    \EndFor
    \end{algorithmic}
    \end{algorithm}

    \subsection{Adaptive Score Fusion}
    \label{sec:fusion}

\begin{figure}[t]
    \centering
    \includegraphics[width=0.9\textwidth]{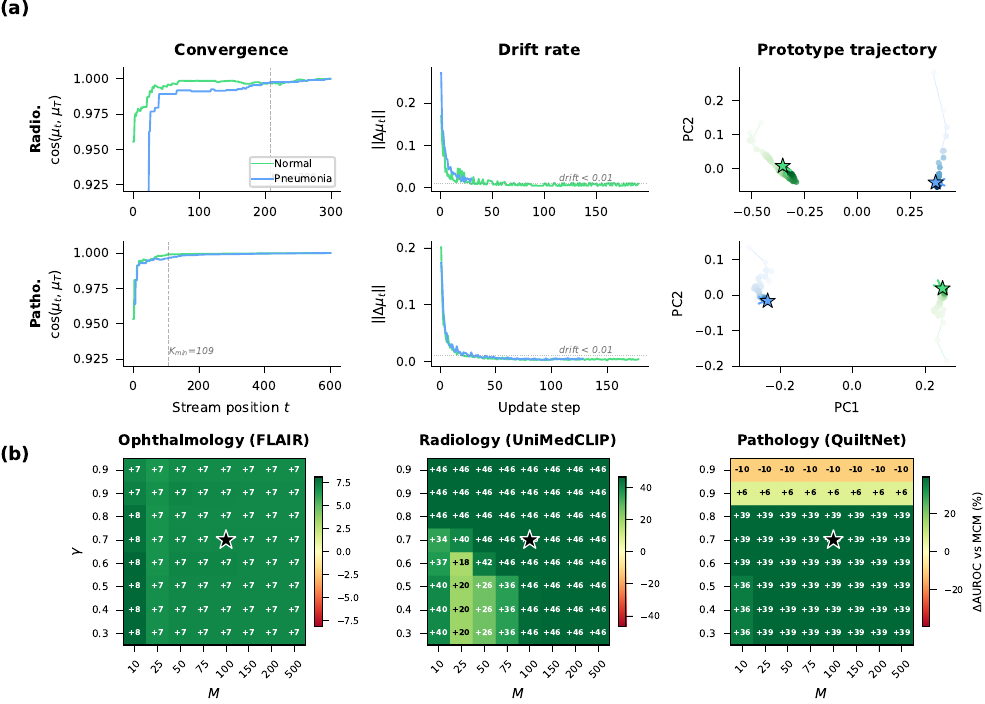}
    \caption{\textbf{PROTON analysis.} \textbf{(a)}~Prototype convergence (cosine similarity to final prototype, drift, and PCA trajectories; $\star$\,=\,final). Dashed lines mark the stream index at which all classes reach $K_{\min}$. \textbf{(b)}~$\gamma \times M$ sensitivity ($\Delta$AUROC over MCM, covariate shift; $\star$\,=\,default) across three modalities.}
    \label{fig:analysis}
\end{figure}

    PROTON fuses both signals into a single score:
    \begin{equation}
        S_{\mathrm{PROTON}}(x_t) = \alpha_t \cdot S_{\mathrm{MCM}}(x_t) + (1 - \alpha_t) \cdot S_{\mathrm{proto}}(x_t),
        \label{eq:fusion}
    \end{equation}
    where $\alpha_t \in [\alpha_{\min}, \alpha_{\max}]$ adapts online rather than requiring per-deployment tuning. We track the running variance $\sigma^2_t$ of $S_{\mathrm{MCM}}$ via Welford's algorithm ($O(1)$ memory, exact single-pass) and set:
    \begin{equation}
        \alpha_t = \alpha_{\max} - \sigma\!\left(100 \cdot (\sigma^2_t - \sigma^2_0)\right) \cdot (\alpha_{\max} - \alpha_{\min}),
        \label{eq:alpha}
    \end{equation}
    where $\sigma(\cdot)$ is the sigmoid and $\sigma^2_0$ a reference threshold corresponding to nominal ID-only variance. Intuitively, a homogeneous (mostly-ID) stream yields low MCM variance, pushing $\alpha_t \to \alpha_{\max}$ (trust MCM); covariate samples produce confidently-wrong softmax that spikes variance, downweighting MCM in favor of $S_{\mathrm{proto}}$. The sigmoid gives smooth, outlier-robust interpolation, and $\alpha_t = 1$ before calibration recovers pure MCM. Under default hyperparameters, PROTON consistently improves over MCM across all shift types (Table~\ref{tab:main}), though extreme gate settings ($\gamma{>}0.9$ or $\gamma{<}0.5$) can degrade performance, as shown in Fig.~\ref{fig:analysis}(b).
    
    \noindent\textbf{Complexity.} PROTON adds negligible overhead: one dot product per class for $S_{\mathrm{proto}}$ ($O(Cd)$), one $O(1)$ queue operation, and three scalar Welford updates per sample. Memory is bounded by $C {\times} M {\times} d$ floats ($2 {\times} 100 {\times} 512 \approx 0.4$\,MB for FLAIR). No gradients, augmentation, or backpropagation are required. Algorithm~\ref{alg:proton} summarizes this.


\section{Experiments}

\noindent\textbf{Benchmark.} We adopt the ophthalmology OOD benchmark of Ju et al.~\cite{ju2025delving}, built on FIVES: ID classes are Normal and Diabetic Retinopathy (150 each); OOD comprises Semantic (Glaucoma + AMD, $N{=}300$), Covariate (UWF fundus, $N{=}150$), and Far-OOD (ImageNet, $N{=}150$). We evaluate on FLAIR~\cite{Silva_Rodr_guez_2025} (ResNet-50, $d{=}512$, Bio\_ClinicalBERT) with frozen weights and vanilla class-name prompts.

\noindent\textbf{Baselines.} We compare against all zero-shot methods from Ju et al.~\cite{ju2025delving} Table~2: Max-Logits, Energy~\cite{liu2020energy}, GL-MCM~\cite{miyai2025gl}, MCM~\cite{ming2022delving}, MCM with hierarchical prompts ($L{=}1,5$), and the prompt-augmentation method TAG-MCM~\cite{liu2024tag}, which like PROTON requires no training data, model modification, or prompt trees. To contextualize the upper bound, we additionally report two \emph{privileged} feature-bank detectors that consume labeled ID features (like kNN-ID and Mahalanobis-ID) which PROTON is not entitled to use. We restrict the main comparison to zero-shot methods because PROTON operates without any labeled data; few-shot approaches (CoOp~\cite{zhou2022cocoop}, LoCoOp~\cite{miyai2023locoop}) occupy a strictly more privileged setting with access to ID training samples, making direct comparison inequitable. We exclude methods that need labeled ID data (HVL~\cite{lai2025hierarchical}, GLAli~\cite{YanJie_Global_MICCAI2025}) or target accuracy rather than detection (BCA+~\cite{li2025generalizing}, SCA~\cite{zhou2025self}, UL-TTA~\cite{kim2025ultra}).

\noindent\textbf{Metrics and hyperparameters.} AUROC, AUPR, and FPR@95 per shift type~\cite{ming2022delving,ju2025delving}. PROTON hyperparameters are fixed across all experiments: $\gamma{=}0.7$, $M{=}100$, $K_{\min}{=}5$, $\alpha_{\min}{=}0.3$, $\alpha_{\max}{=}0.7$, $\sigma^2_0{=}0.02$. The reference variance $\sigma^2_0$ is \emph{not} tuned per-VLM: it is read off the early-stream ID variance from a brief warm-up, and sweeping $\sigma^2_0 \in [0.005, 0.08]$ leaves mean AUROC flat (69.8--71.2) due to sigmoid saturation. We analyze sensitivity to $\gamma$ and $M$ in Sec.~\ref{sec:results}.


\begin{table}[!htb]
\centering
\tiny
\setlength{\tabcolsep}{4pt}
\renewcommand{\arraystretch}{0.8}
\caption{\textbf{Robustness and cross-domain transfer.} \textbf{(a)}~Per-shift AUROC vs.\ stream ID-fraction $\rho$. \textbf{(b)}~Absolute covariate AUROC across three VLMs/modalities (Radiology: Normal/Pneumonia, RSNA, 300 ea.; Pathology: Benign/Malignant, PCam-derived, 300 ea.).}
\label{tab:robust}
\resizebox{\linewidth}{!}{%
\begin{tabular}{@{}l ccc @{\hspace{16pt}} l ccc@{}}
\toprule
\multicolumn{4}{@{}c@{\hspace{12pt}}}{\textbf{(a)} Stream ID-fraction $\rho$}
& \multicolumn{4}{c@{}}{\textbf{(b)} Cross-domain (Cov.\ AUROC)} \\
$\rho$ & Sem & Cov & Far & Domain (VLM) & MCM & PROTON & $\Delta$ \\
\midrule
90\%            & 63.2 & 67.4 & 84.8 & Ophth.\ (FLAIR)        & 42.4 & 66.3 & {\color{deltagreen}+23.9} \\
70\% (default)  & 62.7 & 66.3 & 84.5 & Radio.\ (UniMedCLIP)   & 44.7 & 90.4 & {\color{deltagreen}+45.7} \\
30\%            & 58.8 & 58.5 & 80.4 & Patho.\ (QuiltNet)     & 45.3 & 84.2 & {\color{deltagreen}+38.9} \\
10\%            & 55.1 & 50.9 & 77.2 &                        &      &      &      \\
MCM (ref.)      & 53.9 & 42.4 & 76.4 &                        &      &      &      \\
\bottomrule
\end{tabular}%
}
\end{table}

\section{Results and Analysis}
\label{sec:results}

PROTON achieves 62.7 / 66.3 / 84.5\% AUROC on semantic / covariate / far-OOD (Table~\ref{tab:main}), improving vanilla MCM by +8.8, +23.9, and +8.1 points respectively---the only zero-shot method to improve all three shifts without hierarchical prompts. The +23.9 covariate gain (42.4\%$\to$66.3\%) confirms the core hypothesis: UWF images produce confident softmax yet lie far from ID prototypes in embedding space. At 84.5\% far-OOD, PROTON exceeds even hierarchical MCM $L{=}5$ (82.4\%), showing that prototype distance complements MCM even where it already performs well.


    \begin{table}[!htb]
    \centering
    \tiny
    \setlength{\tabcolsep}{2.8pt}
    \renewcommand{\arraystretch}{0.8}
    \caption{Zero-shot OOD detection on FLAIR\,+\,FIVES. $L$: hierarchical prompt levels. $\Delta$: gain over vanilla MCM.}
    \label{tab:main}
    \resizebox{\linewidth}{!}{%
    \begin{tabular}{@{}ll ccc ccc ccc@{}}
    \toprule
    & & \multicolumn{3}{c}{\textbf{Semantic}} 
      & \multicolumn{3}{c}{\textbf{Covariate}} 
      & \multicolumn{3}{c}{\textbf{Far-OOD}} \\
    \cmidrule(lr){3-5}\cmidrule(lr){6-8}\cmidrule(lr){9-11}
    & Method 
      & AUC$\uparrow$ & APR$\uparrow$ & FPR$\downarrow$ 
      & AUC$\uparrow$ & APR$\uparrow$ & FPR$\downarrow$ 
      & AUC$\uparrow$ & APR$\uparrow$ & FPR$\downarrow$ \\
    \midrule
    \multirow{4}{*}{\rotatebox[origin=c]{90}{\scriptsize\textit{Static}}}
    & Max-Logits                        
      & 40.3 & 35.0 & 97.0 
      & 42.8 & 30.0 & 94.2 
      & 73.8 & 59.2 & 58.2 \\
    & Energy~\cite{liu2020energy}       
      & 39.8 & 35.0 & 97.0 
      & 43.8 & 30.0 & 93.1 
      & 71.4 & 56.6 & 61.9 \\
    & GL-MCM~\cite{miyai2025gl}         
      & 52.1 & 47.6 & 92.4 
      & 43.5 & 30.0 & 93.4 
      & 74.1 & 59.5 & 57.8 \\
    & MCM~\cite{ming2022delving}        
      & 53.9 & 49.7 & 90.1 
      & 42.4 & 30.0 & 94.6 
      & \cellcolor{bronze}76.4 & \cellcolor{bronze}61.9 & \cellcolor{bronze}54.2 \\
    \midrule
    \multirow{2}{*}{\rotatebox[origin=c]{90}{\scriptsize\textit{Hr.}}}
    & MCM ($L{=}1$)                     
      & \cellcolor{bronze}61.6 & \cellcolor{bronze}58.4 & \cellcolor{bronze}80.4
      & \cellcolor{bronze}65.6 & \cellcolor{bronze}48.2 & \cellcolor{bronze}70.0 
      & 52.2 & 36.3 & 91.6 \\
    & MCM ($L{=}5$)$^\dag$
      & \cellcolor{gold}66.7 & \cellcolor{gold}63.6 & \cellcolor{gold}74.0 
      & \cellcolor{gold}87.7 & \cellcolor{gold}70.4 & \cellcolor{gold}46.7 
      & \cellcolor{silver}82.4 & \cellcolor{silver}68.3 & \cellcolor{silver}44.9 \\
    \midrule
    & \cellcolor{protonrow}\textbf{PROTON (Ours)} 
      & \cellcolor{silver}\textbf{62.7} & \cellcolor{silver}\textbf{59.6} & \cellcolor{silver}\textbf{79.0}
      & \cellcolor{silver}\textbf{66.3} & \cellcolor{silver}\textbf{49.0} & \cellcolor{silver}\textbf{69.3}
      & \cellcolor{gold}\textbf{84.5} & \cellcolor{gold}\textbf{70.5} & \cellcolor{gold}\textbf{41.7} \\[1pt]
    & \cellcolor{protonrow}\textit{\quad$\Delta$ vs MCM}
      & \cellcolor{protonrow}{\color{deltagreen}\tiny$\blacktriangle$\,+8.8} 
      & \cellcolor{protonrow}{\color{deltagreen}\tiny$\blacktriangle$\,+9.9}
      & \cellcolor{protonrow}{\color{deltagreen}\tiny$\blacktriangledown$\,11.1}
      & \cellcolor{protonrow}{\color{deltagreen}\tiny$\blacktriangle$\,+23.9} 
      & \cellcolor{protonrow}{\color{deltagreen}\tiny$\blacktriangle$\,+19.0}
      & \cellcolor{protonrow}{\color{deltagreen}\tiny$\blacktriangledown$\,25.3}
      & \cellcolor{protonrow}{\color{deltagreen}\tiny$\blacktriangle$\,+8.1}
      & \cellcolor{protonrow}{\color{deltagreen}\tiny$\blacktriangle$\,+8.6}
      & \cellcolor{protonrow}{\color{deltagreen}\tiny$\blacktriangledown$\,12.5} \\
    \bottomrule
    \end{tabular}%
    }
    {\raggedright\scriptsize $^\dag$Requires domain-expert + ChatGPT-generated hierarchical prompt trees.}
    \end{table}

    \begin{table}[!htb]
    \centering
    \tiny
    \caption{Ablation on the ophthalmology benchmark. \textbf{(a)}~Effect of each PROTON component (AUROC across all shift types). \textbf{(b)}~Sensitivity to confidence threshold $\gamma$ (covariate shift, $M{=}100$). \textbf{(c)}~Sensitivity to bank size $M$ (at $\gamma{=}0.7$).}
    \label{tab:ablations}
    \setlength{\tabcolsep}{2pt}
    \renewcommand{\arraystretch}{1}
    \resizebox{\linewidth}{!}{%
    \begin{tabular}{@{}l ccc @{\hspace{18pt}} l ccc @{\hspace{18pt}} l ccc@{}}
    \toprule
    \multicolumn{4}{@{}c@{\hspace{12pt}}}{\textbf{(a)} Components}
    & \multicolumn{4}{c@{\hspace{12pt}}}{\textbf{(b)} Threshold $\gamma$}
    & \multicolumn{4}{c@{}}{\textbf{(c)} Bank size $M$} \\
    Variant & Cov & Sem & Far
    & $\gamma$ & AUC & APR & F95
    & $M$ & AUC & APR & F95 \\
    \midrule
    MCM baseline           & 42.4 & 53.9 & 76.4  & 0.5  & 66.2 & 48.9 & 69.4  & 10   & 67.1 & 49.6 & 68.4 \\
    +\,Proto.\,($\alpha{=}.5$) & 57.9 & 59.2 & 80.9  & 0.6  & 66.3 & 49.0 & 69.3  & 25   & 66.1 & 48.9 & 69.5 \\
    +\,Adapt.\,fusion      & 62.7 & 61.4 & 82.9  & \cellcolor{protonrow}\textbf{0.7}  & \cellcolor{protonrow}\textbf{66.3} & \cellcolor{protonrow}\textbf{49.0} & \cellcolor{protonrow}\textbf{69.3}  & 50   & 66.4 & 49.1 & 69.2 \\
    \rowcolor{protonrow}
    +\,Conf.\,gate\,(\textbf{Full}) & \textbf{66.3} & \textbf{62.7} & \textbf{84.5}  & 0.8  & 66.2 & 48.9 & 69.4  & \textbf{100}  & \textbf{66.3} & \textbf{49.0} & \textbf{69.3} \\
                           &      &      &       & 0.9  & 65.5 & 48.4 & 70.3  & 200  & 66.2 & 48.9 & 69.4 \\
    \bottomrule
    \end{tabular}%
    }
    \end{table}

\noindent\textbf{Why does PROTON work?} Fig.~\ref{fig:intro}(c) provides a direct answer: t-SNE projections show that ID classes form tight clusters while UWF covariate samples occupy geometrically distinct regions---despite producing nearly identical MCM scores (Fig.~\ref{fig:intro}(a), 88\% overlap). PROTON's prototype bank captures exactly this structure. Fig.~\ref{fig:analysis}(a) confirms that prototypes converge rapidly on the primary ophthalmology benchmark: cosine similarity to the final FLAIR centroids exceeds 0.99 within ${\sim}55$ samples and drift falls below 0.01 after ${\sim}80$ updates, requiring no dedicated calibration phase, with radiology and pathology following the same trajectory. 

\noindent\textbf{Adaptive fusion behavior.} The variance-driven weight $\alpha_t$ responds to stream composition automatically: it drops sharply after covariate-shifted samples enter (MCM variance spikes, upweighting $S_{\mathrm{proto}}$) but remains high for far-OOD where MCM is already discriminative---demonstrating shift-dependent adaptation from statistics alone, without any shift-type labels.

\noindent\textbf{Robustness to stream composition and classifier accuracy.} PROTON assumes a non-zero ID fraction to seed prototypes; we make this explicit and stress-test it. Table~\ref{tab:robust}(a) varies the stream ID-fraction $\rho$: PROTON stays above MCM on all shifts down to $\rho{=}10\%$, because the confidence gate plus the $K_{\min}$ pure-MCM fallback (Alg.~\ref{alg:proton}) keep prototypes uncontaminated and recover the baseline in the worst case, so PROTON never degrades below MCM. Under forced misrouting, it remains above MCM through 30\% label errors (65.5 vs.\ 57.6 mean covariate AUROC). Results are order-stable: over five stream-order permutations the per-shift AUROC is $62.7{\pm}1.1$ / $66.3{\pm}1.8$ / $84.5{\pm}0.9$. Table~\ref{tab:robust}(b) reports absolute covariate AUROC across three VLMs, replacing the earlier $\Delta$-only heatmap: gains are largest on covariate shift everywhere (+23.9 / +45.7 / +38.9), and semantic and far-OOD also improve in all domains.

\noindent\textbf{Ablations.} Table~\ref{tab:ablations} decomposes PROTON's gains. Prototype distance with fixed fusion ($\alpha{=}0.5$) provides the largest single improvement (+15.5 covariate AUROC); adaptive fusion adds +4.8 by dynamically reweighting across shift types; and the confidence gate contributes a further +3.6 by preventing OOD contamination of prototypes. Hyperparameters are robust: performance is stable across $\gamma \in [0.5, 0.8]$ and $M \in [25, 200]$, degrading only at extremes ($\gamma{>}0.9$ starves the bank; $\gamma{<}0.5$ admits OOD noise); Fig.~\ref{fig:analysis}(b) shows this stability holds across the $\gamma \times M$ grid for radiology and pathology too.

\noindent\textbf{Generalization beyond MCM.} PROTON requires only a scalar base score and an embedding, so MCM is one instantiation rather than a dependency. Table~\ref{tab:basescores}(a) instantiates PROTON on six base scores: every one improves on covariate shift (mean $+11$--$15$ AUROC), with semantic and far-OOD following the same pattern, confirming that the gain stems from image-embedding geometry that complements any softmax/logit score. Table~\ref{tab:basescores}(b) further shows PROTON (66.3) matches privileged feature-bank detectors that consume labeled ID features (kNN-ID 70.8, Mahalanobis-ID 68.6) while using none, and composes with hierarchical prompts (Hier.MCM\,+\,PROTON reaches 88.4, exceeding either alone).


\section{Discussion}

    \noindent\textbf{Asymmetric gains and complementarity.} The covariate gain dwarfs the semantic and far-OOD ones because it is the failure mode PROTON targets: UWF images depict the same pathologies as standard fundus (high softmax confidence that blinds MCM) yet differ visually---wider field, different illumination, peripheral structures---placing them in separable embedding regions (Fig.~\ref{fig:intro}(c)) that early-stream prototypes capture. For semantic and far-OOD shifts MCM already separates well, so prototype distance acts as a complement rather than the primary signal.

    Hierarchical MCM ($L{=}5$) achieves a higher covariate AUROC of 87.7\% by enriching \emph{what} the model represents through multi-level clinical text embeddings, whereas PROTON improves \emph{how} OOD scores are derived from existing representations. These two strategies are complementary, not competing: combining them (Hier.MCM\,+\,PROTON) reaches 88.4\% covariate AUROC, above either alone, and PROTON already exceeds Hier.MCM on far-OOD (84.5 vs.\ 82.4); unlike hierarchical prompting, PROTON requires no domain expertise or LLM-generated prompt trees and is deployable out of the box with vanilla class names.
    
    \noindent\textbf{Scope and future work.} PROTON's closed-form fusion could be replaced by a learned gating mechanism for further gains; we defer this to preserve the parameter-free design. Beyond binary detection, the two-score decomposition already supports shift-type \emph{triage}: classifying each sample as covariate/semantic/far-OOD directly from $(S_{\mathrm{MCM}}, S_{\mathrm{proto}})$ reaches macro-F1 $=0.70$ on the 3-way task---high $S_{\mathrm{proto}}$ with low $S_{\mathrm{MCM}}$ flags covariate shift for re-acquisition, while high $S_{\mathrm{MCM}}$ routes to specialist review or rejection. We will extend this triage analysis to additional modalities beyond the three studied here.
    

\begin{table}[t]
\centering
\tiny
\setlength{\tabcolsep}{4pt}
\renewcommand{\arraystretch}{0.7}
\caption{\textbf{PROTON is not MCM-specific.} \textbf{(a)}~Covariate AUROC of six base scores alone (Base) vs.\ as PROTON's base signal (+PROTON). \textbf{(b)}~Covariate AUROC vs.\ broader baselines. $^\ddag$privileged (labeled ID features); $^\dag$LLM-generated prompt trees.}
\label{tab:basescores}
\resizebox{\linewidth}{!}{%
\begin{tabular}{@{}l cc c @{\hspace{16pt}} l c@{}}
\toprule
\multicolumn{4}{@{}c@{\hspace{12pt}}}{\textbf{(a)} PROTON across base scores}
& \multicolumn{2}{c@{}}{\textbf{(b)} Broader baselines} \\
Base score & Base & +PROTON & $\Delta$ & Method & Cov.\ AUC \\
\midrule
Max-Logits          & 42.8 & 62.5 & {\color{deltagreen}+19.7} & TAG-MCM~\cite{liu2024tag}        & 45.7 \\
Energy              & 43.8 & 62.0 & {\color{deltagreen}+18.2} & kNN-ID$^\ddag$               & 70.8 \\
GL-MCM              & 43.5 & 65.8 & {\color{deltagreen}+22.3} & Mahalanobis-ID$^\ddag$       & 68.6 \\
MCM                 & 42.4 & 66.3 & {\color{deltagreen}+23.9} & Hier.MCM ($L{=}5$)$^\dag$    & 87.7 \\
Entropy             & 44.1 & 63.1 & {\color{deltagreen}+19.0} & \textbf{PROTON (ours)}       & \textbf{66.3} \\
TAG-MCM             & 45.7 & 67.1 & {\color{deltagreen}+21.4} & Hier.MCM\,+\,PROTON          & \textbf{88.4} \\
\bottomrule
\end{tabular}%
}
\end{table}

\section{Conclusion}

We presented PROTON, the first test-time adaptive OOD detection framework for medical VLMs. By fusing online embedding-space prototypes with MCM via stream-level variance, it lifts chance-level covariate detection while also improving semantic and far-OOD \textemdash training-free, prompt-free, and parameter-free at 0.4\,MB. More broadly, PROTON demonstrates that the test stream itself contains sufficient structure to improve detection through online prototypes, encouraging further work toward multi-domain validation, shift-type triage, and integration with hierarchical prompting.

\bibliographystyle{splncs04}
\bibliography{reference}

\end{document}